\documentclass[conference]{IEEEtran}
\IEEEoverridecommandlockouts

\usepackage{cite}
\usepackage{amsmath,amssymb,amsfonts}
\usepackage{algorithmic}
\usepackage{graphicx}
\usepackage{textcomp}
\usepackage{xcolor}
\usepackage{placeins}  % Allows for better control of figure placement

\def\BibTeX{{\rm B\kern-.05em{\sc i\kern-.025em b}\kern-.08em
    T\kern-.1667em\lower.7ex\hbox{E}\kern-.125emX}}
\begin{document}

\title{Model Predictive Path Integral Docking of Fully Actuated Surface Vessel}

\author{\IEEEauthorblockN{1\textsuperscript{st} Akash Vijayakumar}
\IEEEauthorblockA{\textit{Department of Ocean Engineering} \\
\textit{Indian Institute of Technology Madras}\\
Chennai, India \\
na21b003@smail.iitm.ac.in}

\and

\IEEEauthorblockN{2\textsuperscript{nd} Atmanand M A}
\IEEEauthorblockA{\textit{Department of Ocean Engineering} \\
\textit{Indian Institute of Technology Madras}\\
Chennai, India  \\
atma@ee.iitm.ac.in}

\and

\IEEEauthorblockN{3\textsuperscript{rd} Abhilash Somayajula}
\IEEEauthorblockA{\textit{Department of Ocean Engineering} \\
\textit{Indian Institute of Technology Madras}\\
Chennai, India  \\
abhilash@iitm.ac.in}

}

\maketitle

% \begin{abstract}
% This paper presents a novel approach to autonomous docking for fully actuated surface vessels using Model Predictive Path Integral (MPPI) control.  We demonstrate a framework that combines LiDAR-based dock detection, real-time trajectory optimization, and robust control for precise docking maneuvers of fully actuated surface vessels. Our method leverages a hybrid cost function incorporating both dock geometry and vessel dynamics, enabling safe and efficient navigation through confined spaces. Simulation results show successful autonomous docking capabilities while maintaining safe clearances and optimal approach trajectories. The MPPI controller generates optimal trajectories by sampling control sequences and evaluating their costs based on wall clearance, orientation alignment, and target position objectives. The effectiveness of the approach is demonstrated through extensive simulation studies using a physics-based model of a fully actuated surface vessel. 
% \end{abstract}

\begin{abstract}
Autonomous docking remains one of the most challenging maneuvers in marine robotics, requiring precise control and robust perception in confined spaces. This paper presents a novel approach integrating Model Predictive Path Integral (MPPI) control with real-time LiDAR-based dock detection for autonomous surface vessel docking. Our framework uniquely combines probabilistic trajectory optimization with a multi-objective cost function that simultaneously considers docking precision, safety constraints, and motion efficiency. The MPPI controller generates optimal trajectories by intelligently sampling control sequences and evaluating their costs based on dynamic clearance requirements, orientation alignment, and target position objectives. We introduce an adaptive dock detection pipeline that processes LiDAR point clouds to extract critical geometric features, enabling real-time updates of docking parameters. The proposed method is extensively validated in a physics-based simulation environment that incorporates realistic sensor noise, vessel dynamics, and environmental constraints. Results demonstrate successful docking from various initial positions while maintaining safe clearances and smooth motion characteristics.
\end{abstract}


% \section{Introduction}
% Autonomous surface vessel operations have gained significant attention in recent years, driven by the increasing demand for automated maritime systems in commercial shipping, research vessels, and recreational boating. Among the various challenges in autonomous vessel navigation, the docking maneuver stands out as particularly complex due to its requirement for precise control in confined spaces, real-time obstacle avoidance, and careful consideration of vessel dynamics.

\section{Introduction}
Autonomous docking of marine vessels represents one of the most challenging maneuvers in maritime autonomy, requiring precise control while accounting for complex vessel dynamics, environmental conditions, and safety constraints. This capability is particularly crucial for unmanned surface vessels (USVs), where reliable autonomous docking enables extended operations and reduced human intervention. The challenge lies not only in achieving the final docked position but also in executing a safe and efficient approach trajectory while maintaining proper clearance from dock structures.

Traditional control approaches for autonomous docking often rely on predetermined trajectories or reactive behaviors, which may not optimally handle the full complexity of the docking maneuver. Model Predictive Control (MPC) offers a more sophisticated approach by continuously optimizing future trajectories, but classical MPC implementations can struggle with the nonlinear dynamics and uncertainty inherent in marine environments. This motivates the exploration of sampling-based MPC methods, particularly the Model Predictive Path Integral (MPPI) control framework \cite{MPPI_theory_to_parallel}, which can naturally handle nonlinear dynamics and non-convex constraints through probabilistic sampling.

In this paper, we present a comprehensive MPPI-based control framework specifically designed for autonomous vessel docking. Our approach incorporates several key innovations:
\begin{itemize}
    \item A multi-objective cost function that balances docking precision, safety constraints, and motion efficiency
    \item Real-time processing of LiDAR data for dock detection and relative positioning for cost computation.
    \item Computation of costs without access to prior map information of the environment. 
\end{itemize}

The control framework is extensively validated in a physics-based simulation environment that captures essential aspects of marine vessel dynamics and sensor characteristics. Our implementation considers a fully actuated surface vessel equipped with omnidirectional thrust capability, similar to modern autonomous water taxis and small marine research platforms. The simulation incorporates realistic sensor noise and precise collision detection to evaluate the robustness of the control strategy.

Results demonstrate the effectiveness of our approach in achieving precise docking maneuvers while maintaining safe clearances and smooth motion characteristics. The controller successfully handles various initial positions and orientations, demonstrating its potential for practical deployment in autonomous marine systems. Quantitative analysis of docking accuracy, path efficiency, and motion smoothness provides insights into the controller's performance characteristics.

% The remainder of this paper is organized as follows: Section II presents the mathematical formulation of the MPPI control framework and our specific cost function design. Section III details the simulation environment used for validation. Section IV presents experimental results and performance analysis. Finally, Section V provides conclusions and discusses potential future developments in autonomous marine docking systems.

\section{References and Related Work}
The task of autonomous vessel docking has been approached through various methodologies in recent literature. Here, we discuss key related works and highlight how our approach differs.

Homburger et al. \cite{HomburgerWirtensohnReuter2022} applied MPPI control to a fully-actuated surface vessel, achieving precise docking maneuvers through a cost function that accounts for docking position, orientation, and actuator dynamics. While similar in using MPPI, our work extends this by incorporating lidar scan instead of a prior map of environment and cost formulation for practical docking scenarios.

Streichenberg et al. \cite{Streichenberg_2023} proposed a decentralized MPPI framework for multi-agent path planning in urban canals, focusing on interaction-aware motion planning. While they demonstrated success in navigation through narrow canals, their work did not specifically address the docking problem that we tackle.

Patel et al. \cite{patel_2023} presented an MPPI framework for USV control under wave disturbances using Froude-Krylov theory. While they showed robust control under environmental disturbances, their work focused on trajectory tracking rather than the specific challenges of docking maneuvers that we address.

% Our approach offers several key advantages over existing methods:
% \begin{itemize}
%     \item Incorporation of lidar information and point cloud segmentation into cost formulation.
%     \item Explicit consideration of safety clearances and dock structure in trajectory planning
%     \item Novel multi-objective cost function balancing docking precision with efficient approach trajectories
%     % \item Real-time computation through efficient MPPI implementation
% \end{itemize}

% However, there are also some limitations in our current approach:
% \begin{itemize}
%     \item Higher computational requirements compared to simpler control methods
%     \item Need for careful tuning of cost function weights
%     \item Assumes relatively accurate vessel state estimation
%     \item Limited handling of extreme environmental conditions
% \end{itemize}

The key contribution of our work lies in bridging the gap between theoretical MPPI formulations and practical autonomous docking implementations by incorporating real-world considerations while maintaining computational feasibility. Future work could address the limitations through adaptive cost weighting schemes and improved environmental disturbance modeling.

\section{Model Predictive Path Integral (MPPI) Algorithm}

Model Predictive Path Integral (MPPI) control is a sampling-based framework for solving Stochastic Optimal Control (SOC) problems in discrete-time dynamical systems. It is particularly effective in scenarios with complex, non-linear dynamics and non-convex cost functions, leveraging forward simulations to compute optimal control sequences.

\subsection{Algorithm Overview}

The MPPI algorithm solves SOC problems for systems governed by:
\begin{equation}
    \mathbf{x}_{t+1} = f(\mathbf{x}_t, \mathbf{u}_t + \mathbf{n}_t),
\end{equation}
where $\mathbf{x}_t$ is the state vector at time $t$, $\mathbf{u}_t$ represents the control input, $\mathbf{n}_t \sim \mathcal{N}(0, \Sigma)$ denotes Gaussian noise, and $f$ is the system's transition function.

At each iteration, $K$ control sequences $\{\mathbf{U}_k\}_{k=1}^K$ are sampled from a Gaussian distribution $\mathcal{N}(\mathbf{u}_t, \nu\Sigma)$ and used to roll out state trajectories over a horizon $T$:
\begin{equation}
    \mathbf{X}_k = \{\mathbf{x}_{t}, f(\mathbf{x}_{t}, \mathbf{u}_{k,t}), \ldots, f(\mathbf{x}_{t+T-1}, \mathbf{u}_{k,T-1})\}.
\end{equation}

For each trajectory, the associated cost $S_k$ is computed:
\begin{equation}
    S_k = \sum_{t=0}^{T-1} c(\mathbf{x}_t, \mathbf{u}_t) + c_T(\mathbf{x}_T),
\end{equation}
where $c(\cdot)$ is the stage cost, and $c_T(\cdot)$ is the terminal cost.

The weights for importance sampling are calculated using:
\begin{equation}
    w_k = \frac{\exp\left(-\frac{1}{\lambda}(S_k - S_{\min})\right)}{\sum_{j=1}^K \exp\left(-\frac{1}{\lambda}(S_j - S_{\min})\right)},
\end{equation}
where $S_{\min}$ is the minimum cost among all sampled trajectories, and $\lambda$ is a temperature parameter.

The optimal control sequence is computed as:
\begin{equation}
    \mathbf{U}^* = \sum_{k=1}^K w_k \mathbf{U}_k,
\end{equation}
and the first control input $\mathbf{u}^*_0$ from $\mathbf{U}^*$ is applied to the system. This process is repeated at each time step.

\subsection{Advantages of MPPI}

Unlike optimization-based approaches that require linearization of dynamics or convex approximations of cost functions, MPPI directly incorporates non-linear and non-convex aspects. Its reliance on parallelizable sampling enables efficient computation, making it suitable for real-time applications. Additionally, MPPI does not require explicit gradient computations, further enhancing its robustness in handling discontinuous or complex cost landscapes.

% \subsection{Implementation Considerations}

% MPPI requires careful selection of parameters such as the number of samples $K$, the covariance matrix $\Sigma$, and the temperature parameter $\lambda$. The method also benefits from parallel processing capabilities, as the forward simulations and cost evaluations for all samples can be performed independently. This parallelization ensures scalability to systems with high-dimensional states and control inputs.

\section{Simulation Environment}
\begin{figure}[htbp]
\centerline{\includegraphics[width=0.8\linewidth]{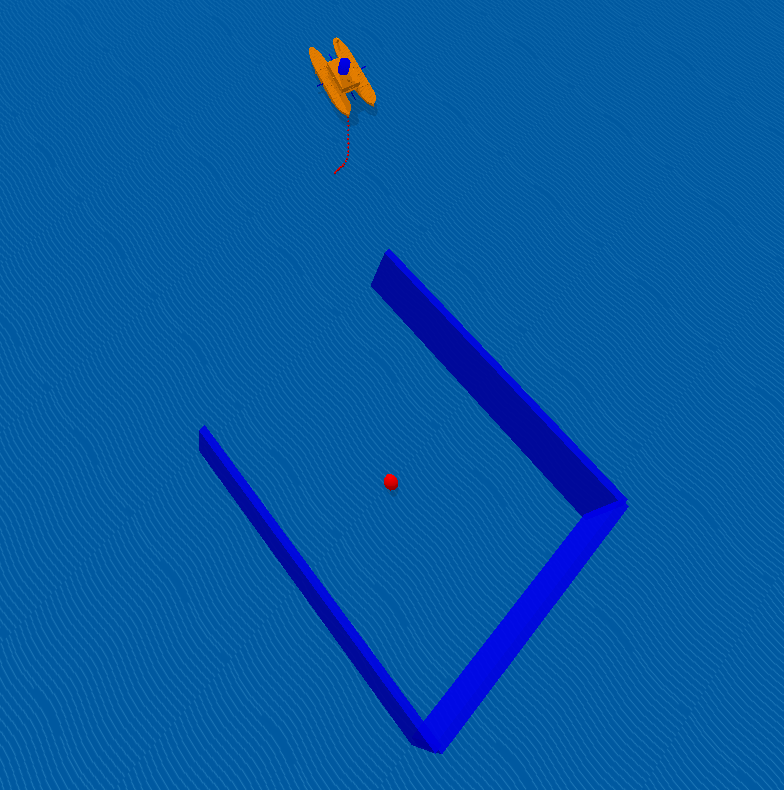}}
\caption{Simulation Environment}
\label{fig}
\end{figure}

The docking simulations were conducted using a PyBullet-based simulation environment, which provides physics-based modeling and visualization capabilities. The environment incorporates several key features designed to replicate realistic vessel docking scenarios.

\subsection{Vessel Dynamics}\label{AA}

The surface vessel is modeled as a fully actuated rigid body operating in 3 degrees of freedom (surge, sway, and yaw). The state vector is defined as:

\begin{equation}
\mathbf{q}_i =
\begin{bmatrix} 
\mathbf{\eta}_i^T & \mathbf{\nu}_i^T 
\end{bmatrix}^T  = 
\begin{bmatrix}
x_i & y_i & \psi_i & u_i & v_i & r_i
\end{bmatrix}^T
\end{equation}

where the state vector is a combination of the 2D pose 

\begin{equation}
\eta = \begin{bmatrix} x & y & \psi \end{bmatrix}^T
\end{equation}

and the velocity vector 

\begin{equation}
\nu = \begin{bmatrix} u & v & r \end{bmatrix}^T
\end{equation}

in body-fixed coordinates.

The dynamic model of the vessel used in running simulations, according to \cite{fossenbook}, is given by:

\begin{equation}
M\dot{\nu} + C_{RB}(\nu)\nu + N\nu = \tau_c + \tau_d
\end{equation}
\begin{equation}
\dot{\eta} = J(\psi)\nu
\label{eq:kinematic}
\end{equation}

The mass matrix \( M \), the Coriolis matrix \( C_{RB}(\nu) \), and the damping matrix \( N \) include the system parameters. The input vector is denoted by \( \tau_c \), and the disturbance vector is denoted by \( \tau_d \). The disturbance vector \( \tau_d \) is set to zero for simplicity for this study. The kinematic equation \eqref{eq:kinematic} describes the transformation of the body-fixed velocity into local coordinates as a function of the rotation matrix \( J(\psi) \). The action input vector has 4 thrust values which helps the vessel to move in all 3 degrees of freedom.

\subsection{Physical Environment}
The simulation environment consists of a water surface rendered with realistic textures and a docking bay structure. The docking bay is modeled with precise physical dimensions:
\begin{itemize}
    \item Dock width: 4 meters
    \item Wall thickness: 0.1 meters
    \item Safety clearance zone: 3.5 meters from center
\end{itemize}

The dock structure includes three primary walls forming a U-shaped berthing area, with wall positions precisely defined relative to the dock center position (10.0, -5.0, 0.0) in the global coordinate frame.

% \subsection{Vessel Modeling}
% The vessel is modeled as a fully actuated surface vehicle with the following characteristics:
% \begin{itemize}
%     \item Length: 1.0 meters
%     \item Width: 0.5 meters
%     % \item Operating height: 0.075 meters above water surface
%     \item Four thrusters providing omnidirectional control
% \end{itemize}

% The vessel's state is represented by a 6-dimensional vector:
% \begin{equation}
%     \mathbf{x} = [x, y, \theta, v_x, v_y, \omega]
% \end{equation}
% where $(x,y)$ is the position, $\theta$ is the heading angle, $(v_x,v_y)$ are linear velocities, and $\omega$ is the angular velocity.

\subsection{Sensor Simulation}
The simulation includes realistic sensor modeling of 2D LiDAR System. A 360-degree LiDAR sensor is simulated with the following specifications:
\begin{itemize}
    \item Angular resolution: 0.1 degrees (3600 rays)
    \item Maximum range: 50 meters
    \item Update rate: 5 Hz
    \item Gaussian noise with $\sigma = 0.1$ meters
\end{itemize}

% \subsubsection{GPS System}
% Position measurements are simulated with added Gaussian noise:
% \begin{equation}
%     \mathbf{p}_{\text{measured}} = \mathbf{p}_{\text{true}} + \mathcal{N}(0, 0.5)
% \end{equation}
% where the standard deviation of 0.5 meters reflects typical GPS accuracy in marine environments.

\subsection{Collision Detection}
The environment implements a collision detection system that monitors:
\begin{itemize}
    \item Contact points between vessel corners and dock walls
    \item Dynamic safety zones with critical (0.25m) and warning (0.5m) thresholds
    \item Real-time collision state feedback for termination conditions
\end{itemize}

% \subsection{Performance Monitoring}
% The simulation includes comprehensive performance monitoring capabilities:
% \begin{itemize}
%     \item Trajectory logging and visualization
%     \item Real-time state recording at 10 Hz
%     \item Docking metrics computation including:
%     \begin{itemize}
%         \item Final position error
%         \item Angular alignment error
%         \item Path efficiency
%         \item Motion smoothness via jerk analysis
%     \end{itemize}
% \end{itemize}

This simulation environment provides a realistic testbed for evaluating the MPPI controller's performance in autonomous vessel docking scenarios, incorporating both the physical complexities of marine operations and practical sensing considerations.
% ~\ref{fig:scenario3}

% \FloatBarrier 

\section{Lidar Scan Processing}

The LiDAR processing pipeline is designed to extract critical geometric features of the dock environment, including the entry point, dock center, orientation, and clearances. These features are subsequently used to guide the vessel's docking maneuvers.

\subsection{Point Cloud Generation}

LiDAR scans provide a set of range and angle measurements. The range data is filtered to exclude points beyond the maximum valid distance, \( d_{\text{max}} \), and the corresponding \((x, y)\) coordinates are computed as:
\begin{equation}
    x = r \cos(\theta), \quad y = r \sin(\theta),
\end{equation}
where \( r \) and \( \theta \) are the range and angle of each LiDAR beam, respectively. The resulting point cloud, \( \mathbf{P} = \{(x_i, y_i)\} \), represents the spatial distribution of objects around the vessel.

\subsection{Clustering and Dock Identification}

The docking station points are extracted using the DBSCAN clustering algorithm \cite{dbscan}. Clusters with sufficient density are retained, and the dock points \( \mathbf{P}_{\text{dock}} \) are identified. The centroid of \( \mathbf{P}_{\text{dock}} \) is calculated to approximate the dock center:
\begin{equation}
    \mathbf{C}_{\text{dock}} = \frac{1}{N} \sum_{i=1}^N \mathbf{p}_i, \quad \mathbf{p}_i \in \mathbf{P}_{\text{dock}},
\end{equation}
where \( N \) is the number of points in \( \mathbf{P}_{\text{dock}} \).

\subsection{Wall Segmentation and Orientation Estimation}

To identify individual dock walls, the Gaussian Mixture Model (GMM) \cite{bishop2006pattern} is applied to segment \( \mathbf{P}_{\text{dock}} \) into \( M \) clusters. For each cluster, a robust line is fitted using RANSAC \cite{ransac}, yielding parameters for the wall as shown in Figure \ref{fig}:
\begin{equation}
    y = mx + c,
\end{equation}
where \( m \) is the slope and \( c \) is the intercept.

Parallel walls are identified by comparing the slopes of the fitted lines. The dock orientation is determined as the average angle of these parallel walls:
\begin{equation}
    \theta_{\text{dock}} = \frac{\arctan(m_1) + \arctan(m_2)}{2}.
\end{equation}

\begin{figure}[htbp]
\centerline{\includegraphics[width=\linewidth]{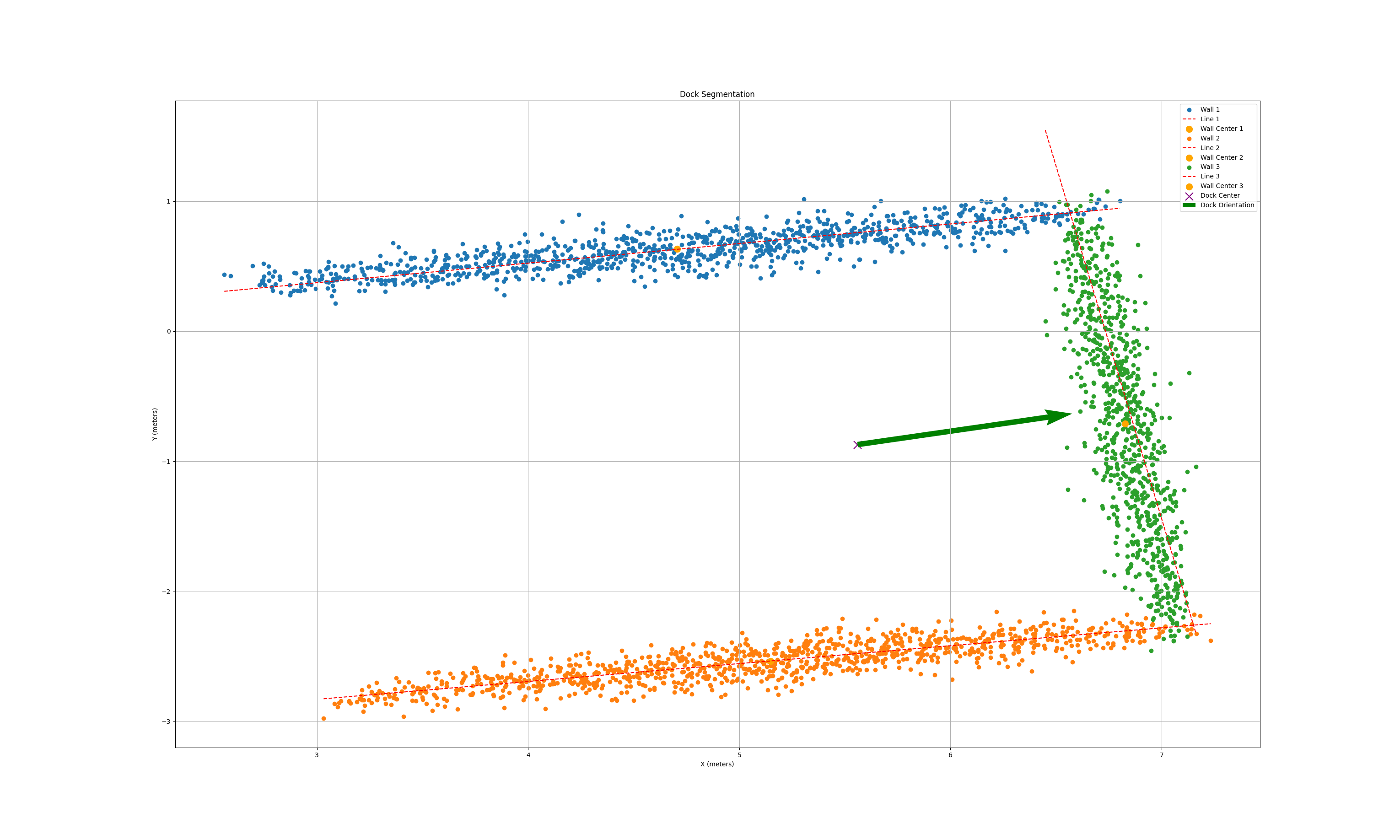}}
\caption{Dock Point Cloud Segmentation}
\label{fig}
\end{figure}

\subsection{Clearance Calculation}

The clearance from the vessel to each wall is computed as the minimum distance from the vessel's corners to the wall:
\begin{equation}
    d_{\text{clearance}} = \min \left( \frac{|m x_0 - y_0 + c|}{\sqrt{m^2 + 1}} \right),
\end{equation}
where \((x_0, y_0)\) are the coordinates of the vessel's corner points.

\subsection{Entry Point Estimation}

The dock entry point is computed as a point offset from the dock center along the dock orientation towards the the open dock face:
\begin{equation}
    \mathbf{P}_{\text{entry}} = \mathbf{C}_{\text{dock}} + d_{\text{entry}} \begin{bmatrix}
        \cos(\theta_{\text{dock}} + \pi) \\
        \sin(\theta_{\text{dock}} + \pi)
    \end{bmatrix},
\end{equation}
where \( d_{\text{entry}} \) is the desired offset distance.

\subsection{Transformation to World Frame}

All calculations are performed in the vessel's local frame. To transform points into the world frame, the rotation and translation induced by the vessel's pose \((x, y, \theta)\) are applied:
\begin{equation}
    \mathbf{P}_{\text{world}} = \mathbf{R} \mathbf{P}_{\text{local}} + \mathbf{t},
\end{equation}
where \( \mathbf{R} \) is the rotation matrix and \( \mathbf{t} \) is the translation vector:
\begin{equation}
    \mathbf{R} = 
    \begin{bmatrix}
        \cos(\theta) & -\sin(\theta) \\
        \sin(\theta) & \cos(\theta)
    \end{bmatrix}, \quad
    \mathbf{t} = 
    \begin{bmatrix}
        x \\
        y
    \end{bmatrix}.
\end{equation}

% The cost formulation is a central aspect of the Model Predictive Path Integral (MPPI) controller, ensuring optimal decision-making by penalizing undesirable states and actions while encouraging desired behaviors. Here it must be designed to ensure precise alignment of the vessel with the dock walls inside the docking station while avoiding collisions with the dock structure. Below, we elaborate on the primary components of the cost function used.

\section{Cost Formulation}
The cost formulation is a central aspect of the Model Predictive Path Integral (MPPI) controller, ensuring optimal decision-making by penalizing undesirable states and actions while encouraging desired behaviors. Below, we elaborate on the primary components of the cost function used.

\subsection{Dock Goal Cost}
The dock goal cost encourages the vessel to move toward the docking position:
\begin{equation}
    C_{\text{dock\_goal}} = w_{\text{dock\_goal}} \cdot \|\mathbf{p} - \mathbf{p}_{\text{dock}}\|_2,
\end{equation}
where $\mathbf{p}$ is the current position of the vessel, $\mathbf{p}_{\text{dock}}$ is the dock center position, and $w_{\text{dock\_goal}}$ is the weight for dock alignment.

\subsection{Velocity Cost}
The velocity cost penalizes deviations from desired motion characteristics in the body frame. The velocity cost formulation is similar to Streichenberg et al\cite{Streichenberg_2023} : 
\begin{equation}
\begin{split}
    C_{\text{velocity}} = & w_{\text{back}} \cdot \text{ReLU}(-v_x) + w_{\text{lat}} \cdot v_y^2 + \\
    & w_{\text{rot}} \cdot \omega^2 + w_{\text{max\_speed}} \cdot (\|\mathbf{v}\| - v_{\text{max}})^2,
\end{split}
\end{equation}
where $v_x$ and $v_y$ are forward and lateral velocities, $\omega$ is the angular velocity, and $v_{\text{max}}$ is the maximum allowed speed of 0.3 m/s.

\subsection{Dock Heading Cost}
This cost ensures proper vessel orientation toward the dock when sufficiently far:
\begin{equation}
    C_{\text{dock\_heading}} = \begin{cases}
        w_{\text{dock\_heading}} \cdot \theta_{\text{err}}^2 & \text{if } d > d_{\text{th}} \\
        0 & \text{otherwise}
    \end{cases}
\end{equation}
where $\theta_{\text{err}}$ is the angle between the vessel's heading and the vector to the dock center, $d$ is the distance to dock, and $d_{\text{th}}$ is the distance threshold.

\subsection{Dock Orientation Cost}
The dock orientation cost enforces proper alignment when near the dock:
\begin{equation}
    C_{\text{goal\_ori}} = \begin{cases}
        w_{\text{goal\_ori}} \cdot (\theta - \theta_{\text{dock}})^2 & \text{if } d < d_{\text{th}} \\
        0 & \text{otherwise}
    \end{cases}
\end{equation}
where $\theta$ is the vessel's heading and $\theta_{\text{dock}}$ is the dock's orientation.

\subsection{Dock Clearance Cost}
To ensure safe operation near the dock walls, a piecewise clearance cost is implemented:
\begin{equation}
    C_{\text{dock\_clear}} = w_{\text{dock\_clear}} \cdot \begin{cases}
        10 & \text{if } d_{\text{min}} < d_{\text{crit}} \\
        5 & \text{if } d_{\text{crit}} \leq d_{\text{min}} < d_{\text{warn}} \\
        0 & \text{otherwise}
    \end{cases}
\end{equation}
where $d_{\text{min}}$ is the minimum distance from vessel corners to dock walls, $d_{\text{crit}}$ is the critical distance threshold of 0.25m, and $d_{\text{warn}}$ is the warning distance threshold of 0.5m.

\subsection{Dock Entrance Cost}
The dock entrance cost guides the vessel through a safe approach trajectory:
\begin{equation}
\begin{split}
    C_{\text{dock\_entrance}} = \begin{cases}
        w_{\text{dock\_entrance}} \cdot \|\mathbf{p} - \mathbf{p}_{\text{ent}}\|_2 & \text{if } \|\mathbf{p}_{\text{curr}} - \\
        & \mathbf{p}_{\text{ent}}\|_2 > 0.5 \\
        0 & \text{otherwise}
    \end{cases}
\end{split}
\end{equation}
where $\mathbf{p}_{\text{ent}}$ is the entrance point position and $\mathbf{p}_{\text{curr}}$ is the current vessel position.

\subsection{Total Cost}
The total cost is computed as the sum of the individual components:
\begin{equation}
\begin{split}
    C_{\text{total}} = & C_{\text{dock\_goal}} + C_{\text{velocity}} + C_{\text{dock\_heading}} + \\
    & C_{\text{goal\_ori}} + C_{\text{dock\_clear}} + C_{\text{dock\_entrance}}.
\end{split}
\end{equation}This formulation balances various objectives, ensuring the vessel achieves precise docking while maintaining safety and efficiency throughout the maneuver.

% \subsection{Total Cost}

% The total cost is computed as the sum of the individual components:
% \begin{equation}
%     C_{\text{total}} = C_{\text{goal}} + C_{\text{velocity}} + C_{\text{heading}} + C_{\text{clearance}} + C_{\text{orientation}} + C_{\text{entrance}}.
% \end{equation}
% This formulation balances various objectives, ensuring the vessel reaches its target efficiently and safely while respecting environmental constraints.

\begin{table}[h!]
\centering
\caption{Cost Parameters Used in the Objective Function}
\label{tab:cost_parameters}
\begin{tabular}{|l|l|l|}
\hline
\textbf{Parameter} & \textbf{Symbol} & \textbf{Value} \\ \hline
Dock Goal Weight & $w_{\text{dock\_goal}}$ & 1.0 \\ \hline
Back Velocity Weight & $w_{\text{back}}$ & 0.08 \\ \hline
Rotational Velocity Weight & $w_{\text{rot}}$ & 10.0 \\ \hline
Lateral Velocity Weight & $w_{\text{lat}}$ & 1.0 \\ \hline
Maximum Speed Weight & $w_{\text{max\_speed}}$ & 5.0 \\ \hline
Within Goal Orientation Weight & $w_{\text{goal\_ori}}$ & 1.0 \\ \hline
Dock Heading to Goal Weight & $w_{\text{dock\_heading}}$ & 3.0 \\ \hline
Dock Clearance Weight & $w_{\text{dock\_clear}}$ & 2.0 \\ \hline
Dock Entrance Weight & $w_{\text{dock\_entrance}}$ & 3.0 \\ \hline
Maximum Speed & $v_{\text{max}}$ & 0.3 m/s \\ \hline
Critical Clearance Threshold & $d_{\text{crit}}$ & 0.25 m \\ \hline
Warning Clearance Threshold & $d_{\text{warn}}$ & 0.5 m \\ \hline
Distance Threshold & $d_{\text{th}}$ & 0.5 m \\ \hline
\end{tabular}
\end{table}

\begin{figure*}[htbp]
\centerline{\includegraphics[width=0.8\textwidth]{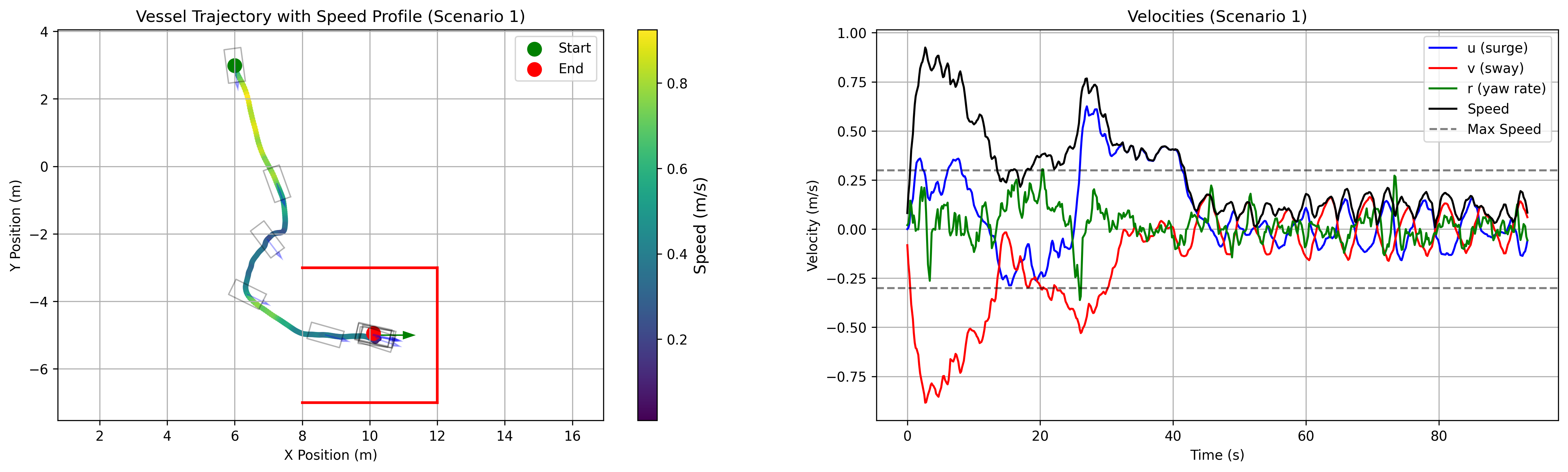}}
\caption{Scenario 1}
\label{fig:scenario1}
\end{figure*}

\begin{figure*}[htbp]
\centerline{\includegraphics[width=0.8\textwidth]{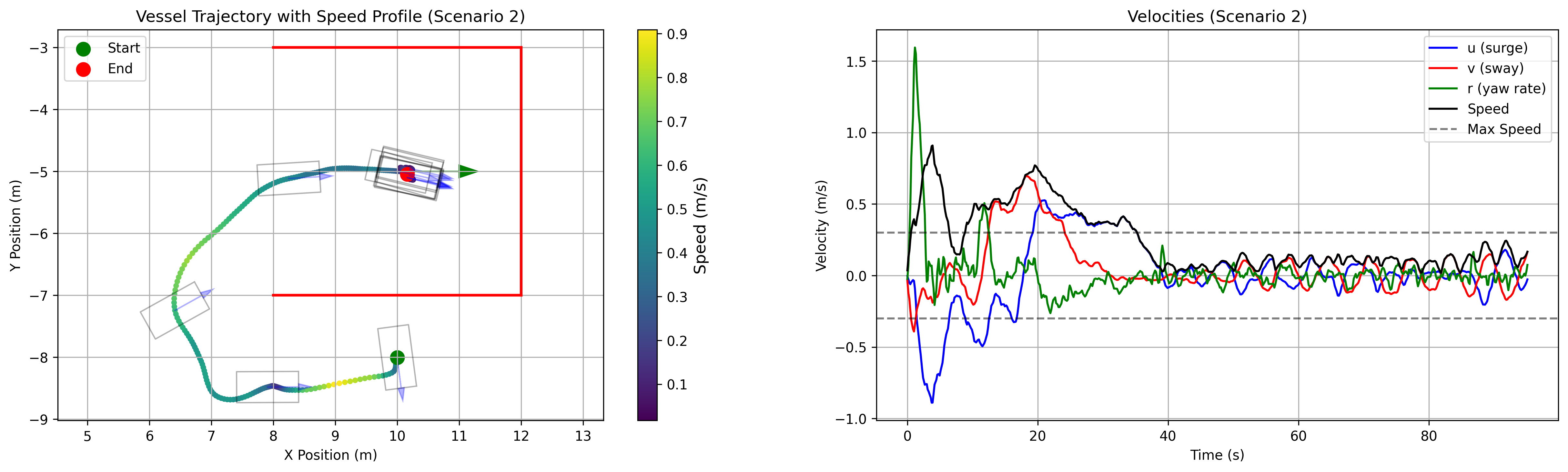}}
\caption{Scenario 2}
\label{fig:scenario2}
\end{figure*}

\begin{figure*}[htbp]
\centerline{\includegraphics[width=0.8\textwidth]{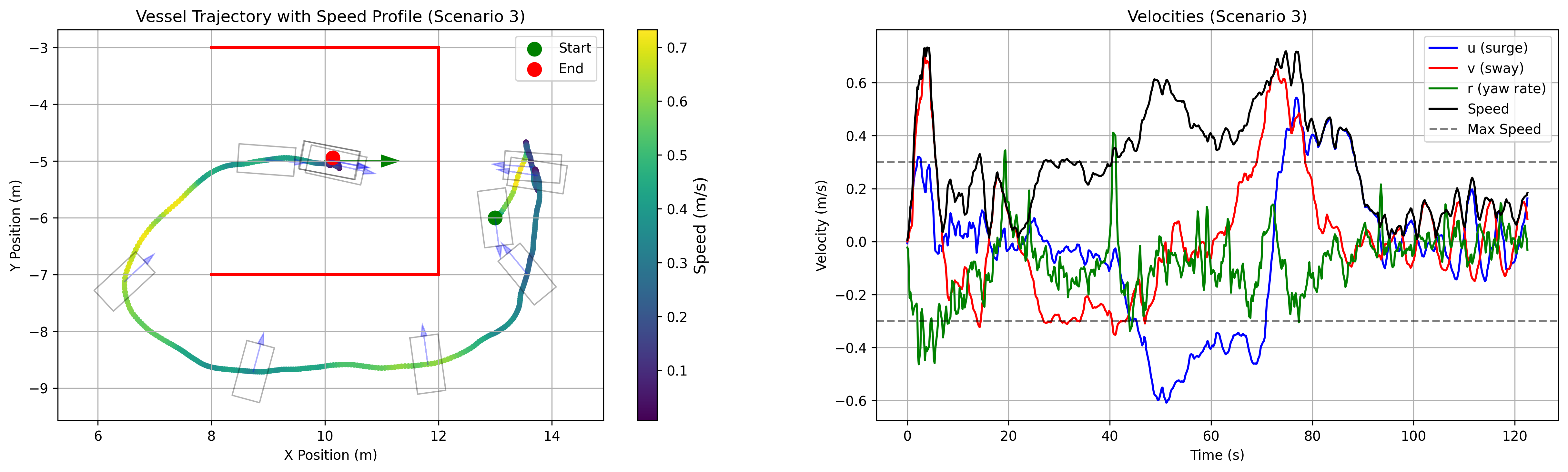}}
\caption{Scenario 3}
\label{fig:scenario3}
\end{figure*}

\begin{figure*}[htbp]
\centerline{\includegraphics[width=0.8\textwidth]{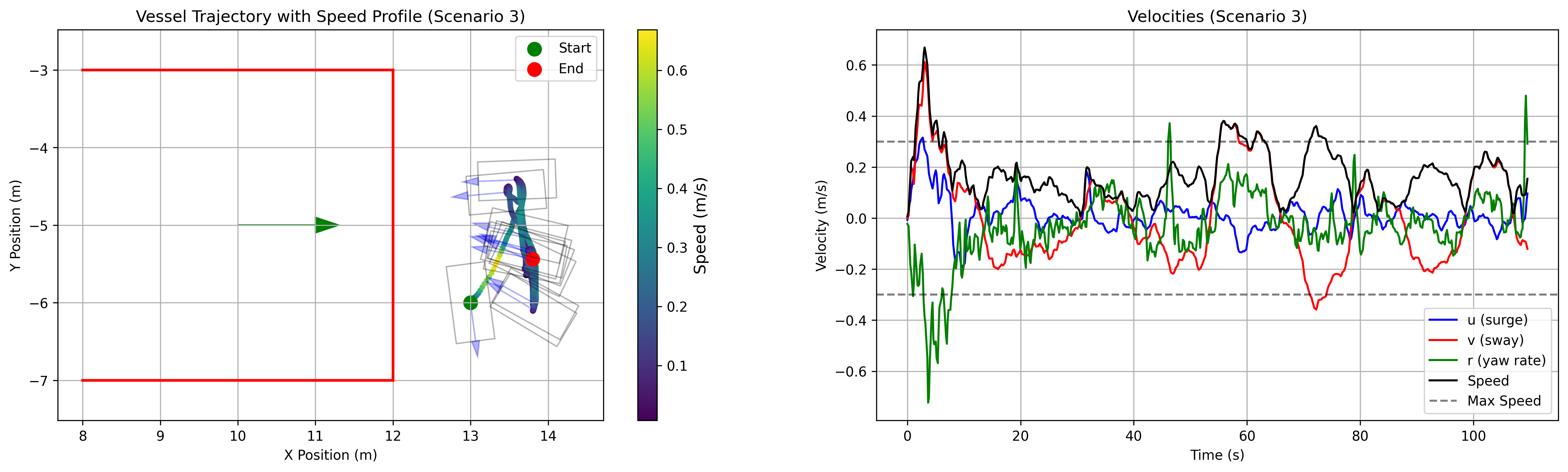}}
\caption{Scenario 3 (Failed)}
\label{fig:scenario4}
\end{figure*}

\section{Results and Discussion}
The proposed MPPI-based docking controller was evaluated through extensive simulation studies under varying initial conditions, with results categorized into three distinct scenarios. In Scenario 1 (Fig. \ref{fig:scenario1}), the vessel begins from a position directly in front of the dock, consistently achieving successful docking maneuvers due to optimal LiDAR visibility of the dock structure. Scenario 2 (Fig. \ref{fig:scenario2}) demonstrates the controller's capability to handle lateral approaches, successfully guiding the vessel from a side position while maintaining safe clearances. Scenario 3 (Fig. \ref{fig:scenario3}) represents the most challenging configuration, with the vessel approaching from behind the docking station, where success rates become variable due to limited dock visibility. Failed attempts (Fig. \ref{fig:scenario4}) primarily occur when approaching from the dock's rear, attributed to poor dock center detection and the stochastic nature of MPPI decision-making under uncertainty. The controller exhibits high reliability in frontal and lateral approaches but shows sensitivity to initialization conditions in scenarios with limited visibility. Notable limitations include small residual velocities persisting after reaching the docking position and unstable behavior observed when the maximum speed weight exceeds 50, occasionally causing the vessel to undock after reaching the target position. These findings highlight the need for refined cost function formulation, particularly in terminal state handling, and suggest potential improvements through enhanced dock detection algorithms or additional sensing modalities for scenarios with limited visibility. Despite these limitations, the controller demonstrates robust performance across most common docking scenarios, effectively balancing safety constraints with docking precision.

% The Model Predictive Path Integral (MPPI) control framework has proven to be a powerful tool for addressing complex stochastic optimal control problems, particularly in the domain of autonomous docking for unmanned surface vessels (USVs). Its ability to handle non-linear dynamics and non-convex cost functions with a sampling-based approach enables precise, collision-free maneuvers in constrained environments.

\section{Conclusion}
This work has demonstrated the effectiveness of Model Predictive Path Integral (MPPI) control for autonomous vessel docking through extensive simulation studies, offering a comprehensive framework that balances docking precision with safety constraints. While our simulation results show promising performance in controlled scenarios, significant challenges remain in transitioning to real-world implementation. These include robust localization in noisy GPS environments, reliable dock detection and segmentation under varying lighting and weather conditions, and accurate estimation of vessel dynamics in the presence of environmental disturbances such as waves, wind, and currents. Our current implementation has identified several areas for improvement, including the need to eliminate residual velocities at the docking position and enhance reliability when approaching from positions with limited dock visibility. Future work should focus on addressing these limitations through adaptive cost function tuning, integration of multiple sensing modalities (LiDAR, vision, radar), and robust point cloud processing algorithms that can handle noisy and incomplete sensor data. Looking to the future, significant advancements can be made by making the weights of the cost function adaptive to dynamic environmental conditions, mission objectives, and vessel states. Furthermore, replacing hand-tuned cost functions with reward functions learned from expert demonstrations via inverse reinforcement learning (IRL) presents an exciting opportunity. By capturing the implicit objectives of human operators, IRL could enable MPPI to generalize across diverse scenarios and reduce the dependency on manual parameter tuning.

\bibliographystyle{IEEEtran}

% \begin{thebibliography}{00}

% \bibitem{b1} G. Eason, B. Noble, and I. N. Sneddon, ``On certain integrals of Lipschitz-Hankel type involving products of Bessel functions,'' Phil. Trans. Roy. Soc. London, vol. A247, pp. 529--551, April 1955.

% \end{thebibliography}

\end{document}